\documentclass[runningheads]{llncs}
\usepackage[T1]{fontenc}
\usepackage{graphicx}
\usepackage{amsmath,amssymb}
\usepackage{booktabs}
\usepackage{longtable,array}
\usepackage{xcolor}
\usepackage[hidelinks]{hyperref}

\usepackage{orcidlink}
\usepackage{marvosym}
\begin{document}
\title{An Ontology-Guided Multi-Anchor Graph Retrieval Framework for Traffic Legal Liability Determination}
\titlerunning{TrafficOmni-RAG}
\author{
Xu Li\inst{1}\textsuperscript{\href{mailto:xul@swpu.edu.cn}{(\Letter)}}\orcidlink{0000-0002-5725-2677}
Shuqi Tian\inst{1}\orcidlink{0009-0000-3674-9898} 
Xun Han\inst{2}\textsuperscript{\href{mailto:hldwxhx@163.com}{(\Letter)}}
Kuncheng Zhao\inst{1}\orcidlink{0009-0000-7448-8043}
Xinyi Li\inst{1}\orcidlink{0009-0008-3346-692X}
}

\authorrunning{Xu Li et al.}
\institute{Southwest Petroleum University, Chengdu, Sichuan, China\\
\email{xul@swpu.edu.cn}, \email{tianshuqi2026@163.com}, \email{zkcmster@163.com},\\
\email{xinyi0733@outlook.com}
\and
Sichuan Police College, Luzhou, Sichuan, China\\
\email{hldwxhx@163.com}
}

\maketitle
\begin{abstract}

Traffic law liability determination is critical for assigning legal penalties, requiring the simultaneous identification of interdependent statutory provisions across multiple legal dimensions. However, existing retrieval-augmented generation methods suffer from a multi-dimensional retrieval bottleneck: single axis architectures compress complex legal queries into a single pathway, causing interdependent statutory dimensions to be overlooked. To address this, we propose OMAGR, an ontology-guided framework that decomposes queries into ontology-aligned anchors and executes parallel graph retrieval across each dimension, ensuring independent retrieval across dimensions before fusion. To evaluate the proposed method, we created the TrafficLaw-QA dataset, an expert-validated benchmark dataset containing 200 questions and 527 legal provisions.
Results show that TrafficOmni-RAG outperforms baselines on \textit{Context Precision} and \textit{Faithfulness} metrics.
The findings demonstrate that parallel multi-anchor retrieval effectively resolves the multi-dimensional retrieval bottleneck, offering a promising direction for traffic law liability determination research.

\keywords{Retrieval Augmented Generation \and Knowledge Graph \and
Multi-Anchor Retrieval \and Traffic Law Liability \and LLM}
\end{abstract}
\section{Introduction}\label{sec:introduction}

Traffic law liability determination is the legal process of assigning responsibility and compensation obligations following a road traffic accident \cite{trafficRAG}. A legally complete determination requires identifying applicable statutes across several interdependent dimensions: the violation committed, the conditions of the accident scene, the roles and statuses of the involved parties, and the rules governing how liability is allocated among them.  Existing deep learning based approaches predict liability labels from accident features without retrieving the underlying statutory provisions\cite{10.1371/journal.pone.0329107}, and ontology-based retrieval systems that search statutes through graph-structured legal ontologies but traverse from a single starting entity\cite{10.1007/978-3-031-36819-6_27}. However, these approaches ignore that a single traffic accident could simultaneously activate provisions governing multiple statutory categories, so an answer confined to one category is legally incomplete.

Retrieval-Augmented Generation (RAG) addresses this structural difficulty by grounding generation in externally retrieved knowledge, enabling dynamic provision retrieval across statutory sources\cite{NEURIPS2020_6b493230}. Existing RAG methods, however, remain constrained by single-axis retrieval architectures. Standard dense retrieval encodes queries and documents into a single vector space, scoring candidates by semantic similarity\cite{karpukhin-etal-2020-dense}. Knowledge graph-augmented RAG systems traverse entity-relation structures from an extracted starting entity\cite{10387715}. Dense retrieval compresses a multi-faceted information need into one query vector, letting the topically dominant dimension crowd out the remaining dimensions. Knowledge graph retrieval begins from a single entity and follows connected edges, limiting traversal to the neighborhood of that entity. For traffic law liability, a single-axis retrieval architecture cannot ensure that every legally relevant dimension contributes evidence. We characterize this as the multi-dimensional retrieval bottleneck.

To address this bottleneck, we propose Ontology-Guided Multi-Anchor Retrieval (OMAGR). OMAGR decomposes an accident description into four anchors aligned with an ontology drawn from a six-dimensional traffic law ontology: accident scene, violation type, responsible party, and liability category. It performs parallel graph retrieval along each anchor simultaneously. Two-hop relational expansion, dense retrieval, and reciprocal rank fusion serve as supporting mechanisms to supplement recall and integrate evidence. Each legally relevant dimension thus receives an independent retrieval pathway before fusion.
We instantiate OMAGR as the retrieval core of TrafficOmni-RAG, an end-to-end legal QA system built on a Neo4j knowledge graph encoding 527 provisions from six Chinese traffic law sources. We evaluate the system on TrafficLaw-QA, a 200-item expert-verified benchmark annotated under a double-blind protocol. Our main contributions include:

\begin{enumerate}
    \item we propose OMAGR to resolve the multi-dimensional retrieval challenge through parallel multi-anchor graph retrieval.
    \item we introduce TrafficLaw-QA for multi-dimensional legal retrieval evaluation.
    \item we demonstrate through experiments and ablation that TrafficOmni-RAG substantially improves \textit{Context Precision} and \textit{Faithfulness} over baselines.
\end{enumerate}

\section{Related Work}\label{sec:related-work}

Standard RAG encodes queries and documents into individual dense vectors scored by their dot product\cite{karpukhin-etal-2020-dense,NEURIPS2020_6b493230}, compressing multi-faceted information needs into one representation. Query rewriting\cite{wang-etal-2023-query2doc} and iterative retrieval\cite{ICLR2024_25f7be96} improve recall diversity, but operate within a single semantic space without restructuring retrieval around the dimensional organization of the domain. In legal QA, this compression is particularly damaging because a single accident simultaneously activates provisions governing different statutory categories (violation, scene, party, liability), yet a single query vector cannot represent all of them simultaneously. Graph RAG\cite{edge2025localglobalgraphrag} introduces entity-centric summarization over community structures yet still follows a single retrieval pathway rather than simultaneous multi-dimensional traversal.

Knowledge graphs have been incorporated into RAG pipelines to inject entity and relation structure into retrieval. In the legal domain, general legal LLMs such as ChatLaw\cite{cui2024chatlawmultiagentcollaborativelegal} and DISC-LawLLM\cite{yue2023disclawllmfinetuninglargelanguage} incorporate knowledge bases for statutory recommendation rather than multi-dimensional liability reasoning; Lawyer LLaMA\cite{huang2023lawyerllamatechnicalreport} similarly treats legal knowledge as a flat article collection. KG-based retrieval systems, exemplified by LlamaIndex KG\cite{9bca3c55c68d44ceaaa4ed350a02ccc0}, traverse from a single starting entity: if the initial extraction misses a legally relevant dimension, the corresponding provisions remain unreachable. TrafficRAG\cite{trafficRAG} applies hybrid BM25-dense retrieval to Chinese traffic law consultation but does not differentiate among legal dimensions in its ranking function. None of these systems restructures retrieval around the simultaneous activation of multiple interdependent provisions that characterizes traffic law liability determination.

Existing legal benchmarks evaluate judgment prediction\cite{xiao2018cail2018largescalelegaldataset}, contract review and statutory interpretation\cite{NEURIPS2023_89e44582}, and language model performance on legal text\cite{chalkidis-etal-2020-legal}, but none isolates the retrieval challenge that arises when a single legal question requires recovering provisions from multiple interdependent statutory dimensions. TrafficLaw-QA addresses this gap with 200 items spanning 11 thematic categories, each designed to require multi-statute grounding, annotated under a double-blind protocol (Cohen's $\kappa = 0.82$). OMAGR restructures retrieval around a six-dimensional legal ontology, querying four anchors in parallel to ensure each legally relevant dimension contributes evidence before fusion.

\section{Methodology}\label{sec:methodology}

\begin{figure}[!htb]
\centering
\includegraphics[width=\textwidth]{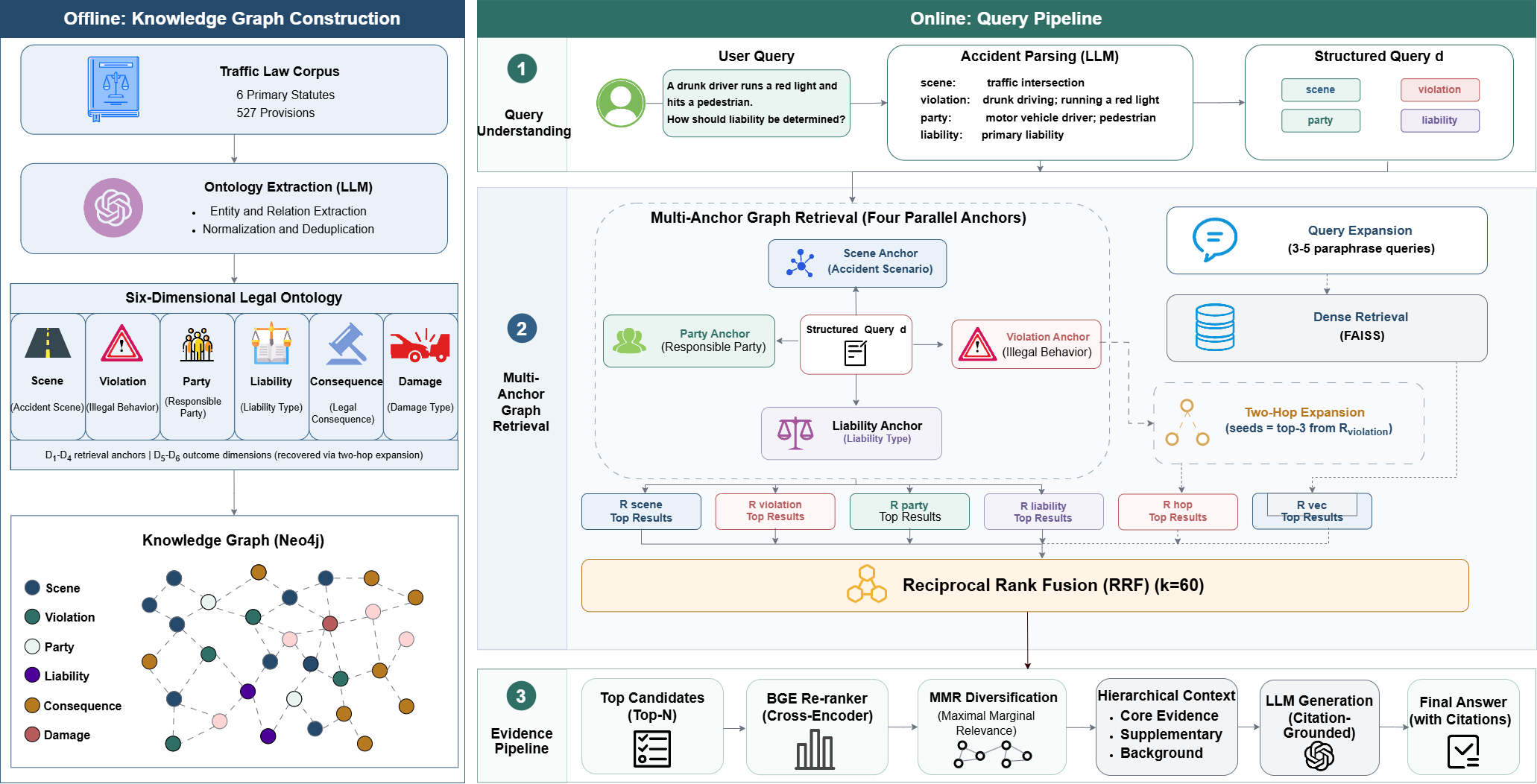}
\caption{TrafficOmni-RAG framework.}\label{fig:architecture}
\end{figure}

TrafficOmni-RAG consists of two stages. In the offline stage (§3.1), we
construct a Neo4j knowledge graph from 527 provisions drawn from six
Chinese traffic law sources, organized under a six-dimensional legal
ontology. In the online stage, the OMAGR framework (§3.2) decomposes
each user query into structured legal anchors, performs parallel graph
retrieval along each anchor, expands results through two-hop relational
traversal, fuses all evidence via reciprocal rank fusion, and refines
the final set; the generator (§3.3) then produces a citation-grounded
answer from the refined evidence. Figure~\ref{fig:architecture} illustrates the full
architecture.

\subsection{Legal Ontology and Knowledge Graph Construction}\label{sec:ontology-kg}

A complete traffic law liability determination requires reasoning across multiple legally interdependent dimensions. The applicable rules depend on the accident scene; the penalty provisions depend on the violation; the rights and obligations depend on the parties; and the liability allocation depends on the interaction among all three. We define a six-dimensional legal ontology $O = \{D_1, ..., D_6\}$ that captures this interdependency by organizing provisions according to their legal function. The six dimensions are accident scene ($D_1$), violation type ($D_2$), responsible party ($D_3$), liability category ($D_4$), legal consequence ($D_5$), and damage type ($D_6$).

\begin{figure}[!htb]
\centering
\includegraphics[width=0.9\textwidth]{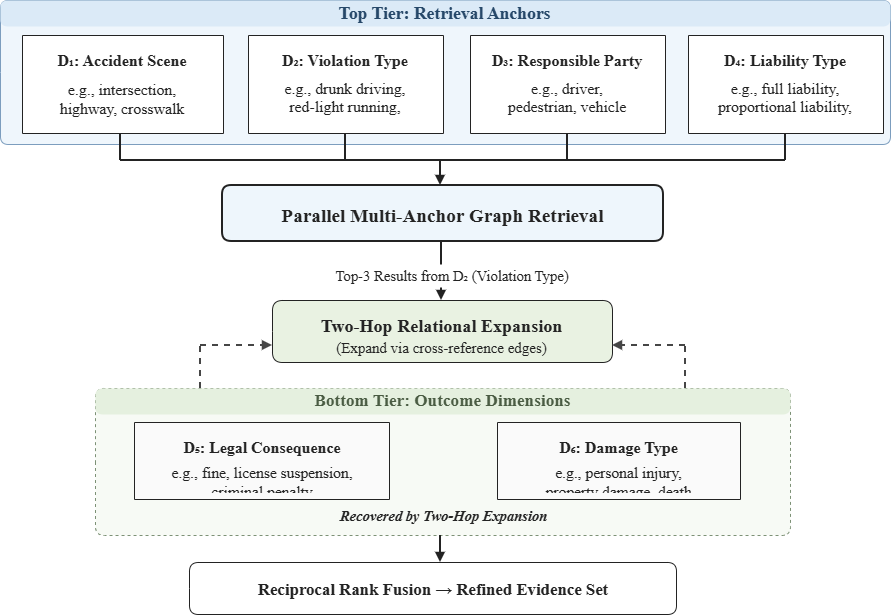}
\caption{Six-dimensional traffic law ontology. The four retrieval anchors $D_1$--$D_4$ (blue) feed parallel graph queries; the two outcome dimensions $D_5$--$D_6$ (green) are recovered through two-hop relational expansion. All six dimensions converge via RRF fusion into the refined evidence set for citation-grounded generation.}\label{fig:ontology}
\end{figure}

Dimensions $D_1$–$D_4$ serve as the four retrieval anchors in the OMAGR framework (§3.2): they characterize the factual and legal premises from which relevant provisions must be retrieved. The scene dimension ($D_1$) identifies location-dependent traffic rules; the violation dimension ($D_2$) captures the legal classification of the infraction; the party dimension ($D_3$) specifies the roles of involved actors; and the liability dimension ($D_4$) governs how legal responsibility is allocated. Dimensions $D_5$–$D_6$ characterize legal outcomes rather than retrieval inputs. Consequence ($D_5$) specifies penalties and sanctions; damage ($D_6$) covers injury categories and property loss types. These two dimensions are not directly queried; they are recovered through the relational expansion step (Algorithm 1, line 7), where cross-references connect violation anchor results to corresponding penalties. Figure~\ref{fig:ontology} illustrates the ontology and its mapping to the OMAGR retrieval framework. 

We map each provision to its applicable ontology dimensions via LLM extraction. A single provision may span multiple dimensions: for instance, Article~91 of the Road Traffic Safety Law (drunk driving penalties) maps to $D_2$ (violation), $D_4$ (liability), and $D_5$ (consequence). The resulting knowledge graph $G=(V,E)$ stores provisions as nodes with typed edges encoding (i)~ontology-dimension associations, (ii)~inter-provision cross-references, and (iii)~statutory hierarchies (law, regulation, judicial interpretation). The graph is implemented in Neo4j, and all provision embeddings are indexed in a FAISS vector store for semantic retrieval.

\subsection{The OMAGR Framework}\label{sec:omagr-framework}

OMAGR is the core retrieval framework. It addresses the multi-dimensional retrieval bottleneck by replacing single-axis retrieval with parallel retrieval across four ontology-aligned anchors. As formalized in Algorithm 1, the pipeline distinguishes between primary and supporting mechanisms: anchor extraction and parallel graph queries constitute the primary retrieval pathway, ensuring that every legally relevant dimension contributes independent evidence prior to fusion. Meanwhile, query expansion, dense retrieval, two-hop relational expansion, reciprocal rank fusion, and evidence refinement operate as supporting mechanisms to enhance recall, mitigate sparsity, and improve the precision of fused results.

Multi-anchor retrieval differs fundamentally from query expansion, and
the two mechanisms operate at different levels. Query expansion
generates diverse reformulations within a single semantic space to
improve lexical and topical recall. Multi-anchor retrieval, in contrast,
decomposes the query along predefined ontology dimensions and executes
independent graph traversals in distinct subgraph regions, each aligned
with a specific legal dimension. The two mechanisms are complementary:
query expansion broadens coverage within each anchor's
retrieval pathway, while multi-anchor retrieval ensures that all legally
relevant dimensions are represented before fusion.

\begin{center}
\rule{\textwidth}{0.4pt}
\vspace{2pt}
{\centering \textbf{Algorithm 1: OMAGR Pipeline}\par}
\vspace{-8pt}
\rule{\textwidth}{0.4pt}\\[4pt]
\begin{minipage}{0.92\textwidth}
\small
\noindent \textbf{Input:} query $q$, KG $G$ (built on ontology $O$), vector index $V$\\[2pt]
\textbf{Output:} diversified provision set $R$, $|R| \leq 8$\\[4pt]
1:\quad $d \gets \textsc{ParseAccident}(q)$ \hfill \\[2pt]
2:\quad $Q \gets \textsc{QueryExpansion}(q, d)$ \hfill \\[4pt]
3:\quad \textbf{for} $k \in \{\text{scene}, \text{violation}, \text{party}, \text{liability}\}$ \textbf{do}\\[2pt]
4:\quad\quad $R_k \gets \textsc{GraphSearch}(G, d_k)$ \hfill \\[2pt]
5:\quad \textbf{end for}\\[4pt]
6:\quad $R_{\text{vec}} \gets \textsc{DenseSearch}(V, Q)$ \hfill \\[2pt]
7:\quad $R_{\text{hop}} \gets \textsc{TwoHopExpand}(G, \text{top3}(R_{\text{violation}}))$\\[4pt]
8:\quad $R \gets \textsc{RRF}(\bigcup_k R_k \cup R_{\text{vec}} \cup R_{\text{hop}},\; k{=}60)$\\[4pt]
9:\quad $R \gets \textsc{BGERerank}(q, R, \text{top}{=}8)$\\[2pt]
10:\ \ $R \gets \textsc{MMR}(R, \lambda{=}0.7)$\\[4pt]
11:\ \ \textbf{return} $R$
\end{minipage}\\[4pt]
\rule{\textwidth}{0.4pt}
\end{center}

Anchor extraction uses constrained LLM parsing with a fixed output schema. If a query provides insufficient information for a particular anchor, that branch is skipped without blocking the pipeline. In parallel, query expansion generates from 3 to 5 paraphrased formulations for dense retrieval diversity\cite{wang-etal-2023-query2doc}. Each anchor then issues a typed Cypher query traversing the ontology edge for that dimension. This design is the central architectural choice: because each anchor queries a distinct region of the graph, the dominant semantic signal (typically the violation) cannot crowd out other dimensions; every legally relevant dimension receives an independent retrieval pathway.

Dense retrieval over the expanded queries provides semantic fallback for provisions lacking strong graph connectivity\cite{karpukhin-etal-2020-dense}. Two-hop expansion traverses inter-provision cross-reference edges from the top 3 violation anchor results, recovering co-applicable statutes not directly reachable through ontology edges (for example, a penalty provision referencing a separate procedural article). All provision sets are merged via Reciprocal Rank Fusion\cite{10.1145/1571941.1572114} ($k=60$) with uniform weights across sources. A cross-encoder reranker\cite{10.1145/3626772.3657878} re-scores and retains the top 8 provisions. Maximal Marginal Relevance\cite{10.1145/290941.291025} removes near-duplicates; and hierarchical context pruning allocates token budget by reranker score, assigning full text to core provisions and progressively shorter excerpts to supplementary ones.

\subsection{Generation}\label{sec:generation}

The refined provision set $R$ ($| R|{}
\leq 8$) forms the retrieval-augmented context for the
generator LLM (GLM-4-Flash). We use citation-grounded generation\cite{gao-etal-2023-enabling}: the model is instructed to embed verbatim statutory phrases as evidence markers and is prohibited from citing provisions outside $R$. This constrains every answer claim to a verifiable statutory source, reducing
hallucination risk in legal reasoning.

\section{TrafficLaw-QA Benchmark}\label{sec:trafficlaw-qa}

Standard legal QA benchmarks evaluate whether models can apply legal knowledge to specific tasks: criminal judgment prediction\cite{xiao2018cail2018largescalelegaldataset}, contract
review\cite{NEURIPS2023_89e44582}, statutory interpretation\cite{chalkidis-etal-2020-legal}. None, however, is designed to expose
the retrieval problem at the center of this paper: cases where legally
complete answers require recovering provisions from multiple
interdependent statutory dimensions simultaneously. TrafficLaw-QA is
constructed to fill this gap. It contains 200 expert-verified items
spanning 11 thematic categories, each designed to require
multi-provision grounding for a legally sufficient answer.

\subsection{Corpus and Scope}\label{sec:corpus-scope}

TrafficLaw-QA draws from six Chinese traffic law sources spanning three tiers of legal authority: the Road Traffic Safety Law (legislation); four administrative regulations, namely the Implementation Regulations of the Road Traffic Safety Law, the Provisions on Handling Road Traffic Violations, the Traffic Accident Handling Procedure Regulations, and the Regulations on Compulsory Insurance for Motor Vehicle Traffic Accident Liability (MTPLI); and the Supreme Court Interpretations on Road Traffic Accident Damage Compensation (judicial interpretation). Together, these sources define the legal framework for traffic accident liability determination in China.

\subsection{Question Design}\label{sec:question-design}

The 200 questions span 11 thematic categories and three difficulty levels (Table~\ref{tab:trafficlaw-qa} and Table~\ref{tab:difficulty}). Categories range from high-frequency liability scenarios such as liability determination and legal penalties to specialized subdomains such as highway rules and traffic signals, reflecting the diversity of real-world traffic law consultation.

Difficulty is defined by the number of statutory provisions required for a legally complete answer. Easy questions require direct application of a single provision. Medium questions require integrating two related provisions. Hard questions require multi-provision reasoning across several legal dimensions. The distribution skews toward Medium and Hard, which together account for 72.5\% benchmark, reflecting the prevalence of multi-provision scenarios in real traffic law consultation. Each reference answer is constructed from official statutory text as a concise legal explanation grounded in explicit provisions.

\begin{table}[!htb]
\centering
\caption{TrafficLaw-QA distribution across thematic categories and difficulty levels.}\label{tab:trafficlaw-qa}
\begin{tabular}{@{}llll@{}}
\toprule
Category & $n$ & \% & Key Legal Dimensions \\
\midrule
Liability Determination & 59 & 29.5\% & Scene, violation, party, liability \\
Legal Penalties & 23 & 11.5\% & Violation type, penalty category \\
Insurance Compensation & 18 & 9.0\% & MTPLI coverage, claim conditions \\
Accident Procedures & 18 & 9.0\% & Reporting, mediation, investigation \\
Traffic Regulations & 17 & 8.5\% & Speed, lane, right-of-way rules \\
Compensation Scope & 13 & 6.5\% & Damages, injury, property loss \\
Vehicle Management & 13 & 6.5\% & Registration, inspection, licensing \\
Pedestrian / NMV Rules & 13 & 6.5\% & Pedestrian, e-bike, non-motorized \\
Comprehensive Cases & 10 & 5.0\% & Multi-violation, multi-party \\
Highway Rules & 9 & 4.5\% & Speed, distance, emergency stops \\
Traffic Signals & 7 & 3.5\% & Signal lights, signs, markings \\
\midrule
\textbf{Total} & \textbf{200} & \textbf{100\%} & \textbf{11 categories $\cdot$ 44 subcategories} \\
\bottomrule
\end{tabular}
\end{table}


\begin{table}[!htb]
\centering
\caption{Difficulty distribution of TrafficLaw-QA items.}\label{tab:difficulty}
\resizebox{\textwidth}{!}{
\begin{tabular}{@{}llll@{}}
\toprule
Difficulty & $n$ & \% & Reasoning Requirement \\
\midrule
Easy (single-statute) & 55 & 27.5\% & Direct statutory application \\
Medium (two-statute) & 82 & 41.0\% & Integration of related provisions \\
Hard (multi-statute/factor) & 63 & 31.5\% & Multi-dimensional legal reasoning \\
\bottomrule
\end{tabular}
}
\end{table}

\subsection{Expert Verification}\label{sec:expert-verification}

All 200 items were independently reviewed by two legally trained
annotators under a double-blind protocol. The resulting inter-annotator agreement was substantial, with a Cohen's $\kappa$ of 0.82. During adjudication, 183 of the 200 items received minor wording or citation format revisions; no item required substantive legal correction. In addition to verifying the reference answers, the annotators independently identified the set of statutory provisions relevant to each question under the same double-blind protocol; these provision sets serve as the ground truth for Context Recall evaluation.

\section{Experiments}\label{sec:experiments}

\subsection{Experimental Settings}\label{sec:experimental-settings}

All experiments use GLM-4-Flash as the generation model. Dense retrieval
uses FAISS\cite{8733051}; graph storage and traversal use Neo4j. All methods share the
same document corpus, embedding model, chunking strategy, and generation
model, with an identical number of retrieved candidates passed to the
generator.

Evaluation uses RAGAS\cite{es-etal-2024-ragas} with GLM-4-Flash as the judge model under
identical prompting. We report four metrics: Faithfulness (factual
grounding of generated answers in the retrieved context), Answer
Relevancy (whether the answer addresses the query), Context Precision
(fraction of retrieved context that is relevant to the query), and
Context Recall (fraction of relevant ground-truth context that is
retrieved).

\subsection{Baselines}\label{sec:baselines}

We compare TrafficOmni-RAG against three baselines, each chosen to isolate a
specific factor in retrieval design.

Naive RAG performs dense vector retrieval with direct LLM generation,
without structured legal knowledge or graph augmentation. It represents
the conventional single-vector retrieval paradigm and establishes a
reference point for how much multi-anchor retrieval improves over
unstructured semantic search.

TrafficRAG\cite{trafficRAG} proposes hybrid BM25-dense retrieval specialized for Chinese
traffic law consultation. Its pipeline combines sparse lexical matching
with dense semantic search, testing whether improved keyword coverage,
without modeling legal dimensional structure, can close the dimensional
coverage gap that OMAGR targets.

LlamaIndex KG\cite{9bca3c55c68d44ceaaa4ed350a02ccc0} performs entity-relation graph traversal over the same Neo4j knowledge graph used by TrafficOmni-RAG. Retrieval starts from a single extracted entity and walks along connected edges. This is the most controlled comparison: same graph infrastructure, embedding model, and generator, differing only in a single-entity rather than a multi-anchor retrieval strategy.

\subsection{Results}\label{sec:main-results}

Table~\ref{tab:main-results} reports the main results on TrafficLaw-QA. The most informative comparison is between TrafficOmni-RAG and LlamaIndex KG. Both use the same Neo4j knowledge graph, embedding model, and generator. The principal structural difference is retrieval strategy: single-entity traversal rather than multi-anchor parallel retrieval; supporting mechanisms contribute marginally, as the ablation study confirms. TrafficOmni-RAG raises Context Precision from 0.722 to 0.915 and Faithfulness from 0.751 to 0.787. The Context Precision gap is the largest between any two methods in Table~\ref{tab:main-results}: single-anchor traversal retrieves provisions clustered around one entity type while missing co-applicable statutes anchored in other ontological dimensions.

\begin{table}[!htb]
\centering
\caption{Performance comparison on TrafficLaw-QA.}\label{tab:main-results}
\resizebox{\textwidth}{!}{
\begin{tabular}{@{}lllll@{}}
\toprule
Method & Faithfulness$\uparrow$ & Answer Relevancy$\uparrow$ & Context Precision$\uparrow$ & Context Recall \\
\midrule
Naive RAG & 0.625 & 0.736 & 0.834 & 0.891 \\
TrafficRAG & 0.716 & 0.845 & 0.884 & \textbf{0.896} \\
LlamaIndex KG & 0.751 & 0.905 & 0.722 & 0.796 \\
\textbf{TrafficOmni-RAG} & \textbf{0.787} & \textbf{0.912} & \textbf{0.915} & 0.885 \\
\bottomrule
\end{tabular}
}
\end{table}

TrafficRAG achieves the highest Context Recall at 0.896, ahead of 0.885 for TrafficOmni-RAG. Its BM25 component catches keyword-matching provisions that dense retrieval alone may miss, producing broader but less precise evidence sets. TrafficOmni-RAG trades this recall margin for precision: retrieval constrained by ontology edges favors legally relevant provisions over topically similar but legally inapplicable ones. In legal QA, where irrelevant context increases hallucination risk, this precision preference is functionally justified.Naive RAG scores the lowest Faithfulness at 0.625, confirming that unstructured semantic retrieval alone is insufficient for multi-provision legal questions.

\subsection{Ablation Study}\label{sec:ablation}

Table~\ref{tab:ablation} reports the ablation results. Each variant removes or modifies
one component of the full pipeline to isolate its contribution.

\begin{table}[!htb]
\centering
\caption{Ablation analysis on TrafficLaw-QA.}\label{tab:ablation}
\begin{tabular}{@{}lllll@{}}
\toprule
Variant & Faith. & AR & CP & CR \\
\midrule
A0: Naive RAG & 0.625 & 0.736 & 0.834 & 0.891 \\
A1: Vector Only (no KG)$^{\dagger}$ & 0.802 & 0.899 & 0.890 & 0.875 \\
A2: w/o Multi-Anchor$^{*}$ & 0.775 & 0.904 & 0.862 & 0.854 \\
A3: w/o Two-Hop$^{\ddagger}$ & 0.785 & 0.910 & 0.913 & 0.870 \\
A4: w/o Query Expansion & 0.792 & 0.910 & 0.913 & 0.881 \\
A5: w/o MMR & 0.781 & 0.916 & 0.912 & 0.884 \\
\textbf{A6: TrafficOmni-RAG} & \textbf{0.787} & \textbf{0.912} & \textbf{0.915} & \textbf{0.885} \\
\bottomrule
\multicolumn{5}{@{}l}{\footnotesize $^{\dagger}$ Dense retrieval only; no knowledge graph or graph queries.}\\
\multicolumn{5}{@{}l}{\footnotesize $^{*}$ Single-anchor graph retrieval (violation anchor only).}\\
\multicolumn{5}{@{}l}{\footnotesize $^{\ddagger}$ No inter-provision edge traversal.}\\
\end{tabular}
\end{table}

Multi-anchor retrieval. Replacing multi-anchor retrieval with a single graph anchor reduces Context Precision from 0.915 to 0.862, the largest component-level drop in Table~\ref{tab:ablation}. Context Recall declines from 0.885 to 0.854, and Faithfulness falls from 0.787 to 0.775. All four metrics decline together. This pattern shows that single-anchor retrieval does not merely retrieve fewer provisions; it retrieves less relevant ones. Restricting graph traversal to one ontological dimension concentrates the candidate set around that dimension while starving the others. This result directly supports the central claim of the paper: traffic law liability questions require simultaneous retrieval across multiple ontological dimensions.

Vector-only retrieval. Variant A1 removes the knowledge graph and relies solely on dense retrieval. Its Faithfulness of 0.802 exceeds that of the full system by 0.015. Its Context Precision of 0.890 and Answer Relevancy of 0.899 trail the corresponding A6 values by 0.025 and 0.013. This pattern reflects a known property of dense retrieval. Semantic similarity concentrates evidence around the dominant topic of the query, typically the violation. The resulting candidate set is precise but dimensionally narrow. In multi-party and multi-violation scenarios, this concentration systematically omits co-applicable statutes from other dimensions. A1 outperforms the single-anchor graph variant, A2, on Context Precision by 0.028. Only the full multi-anchor system combines the precision of semantic retrieval with the dimensional coverage of parallel graph traversal.

Relational expansion. Variant A3 removes two-hop relational expansion. Its metrics are nearly identical to those of A6. Faithfulness, Answer Relevancy, and Context Precision each differ by no more than 0.002. Context Recall drops by 0.015. The negligible gap on the first three metrics indicates that two-hop traversal adds little beyond what the four anchor pathways and dense retrieval already supply. The modest recall gain aligns with the design purpose of relational expansion. It traverses inter-provision cross-reference edges to surface co-applicable statutes that are absent from direct ontology queries. These additional provisions widen the candidate pool without degrading precision, but their impact on final answer quality remains marginal.

Supporting mechanisms. Query expansion, corresponding to A4, shifts each metric by at most 0.005. Two-hop relational expansion, A3, shifts Context Precision and Faithfulness by no more than 0.002, with a Context Recall decline of 0.015. MMR diversification, A5, shifts each metric by at most 0.006. Across all supporting mechanisms, Context Precision varies by no more than 0.003 from the full system. The principal performance gains originate from the multi-anchor retrieval structure.

\section{Analysis}\label{sec:analysis}

\subsection{Retrieval Source Diversity and Dimensional Coverage}\label{sec:dimensional-coverage}

\begin{figure}[!htb]
\centering
\includegraphics[width=0.8\textwidth]{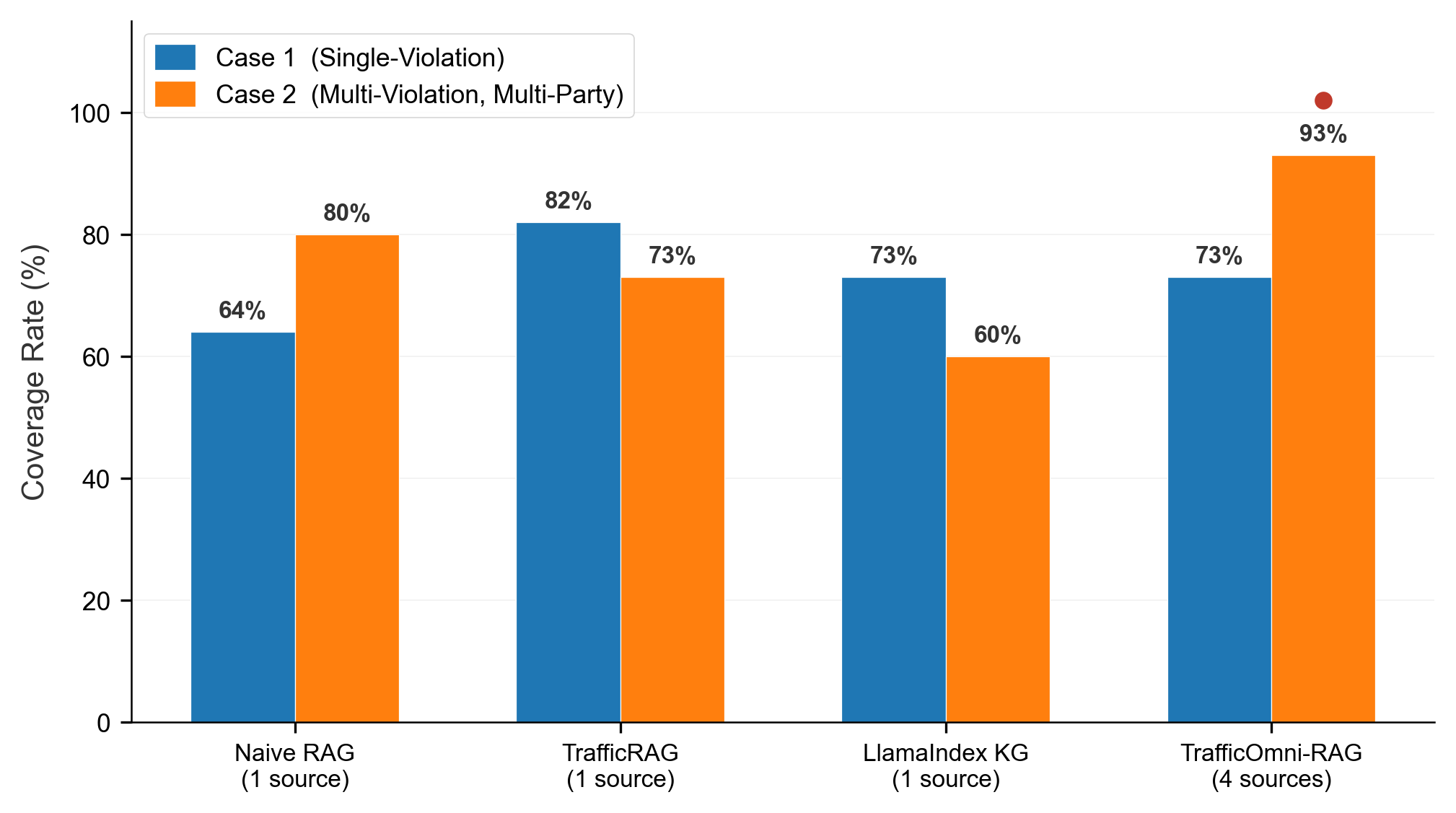}
\caption{Per-dimension retrieval coverage on two representative cases. Case~1 (single-violation, multi-consequence): a driver runs a red light while intoxicated, strikes a pedestrian at a crosswalk, and flees; 11 dimension values are expected. Case~2 (multi-violation, multi-party): an unlicensed driver runs a red light, causing a chain collision involving three vehicles with expired insurance; 15 dimension values are expected. Retrieval source count is shown below each method. TrafficOmni-RAG is the only method drawing from four retrieval sources; the red marker highlights its 93\% coverage on Case~2.}\label{fig:coverage}
\end{figure}

OMAGR retrieves from four distinct sources (scene, violation, party, and liability) while each baseline draws from a single source. Figure~\ref{fig:coverage} reports per-dimension retrieval coverage on two representative cases; a dimension value is covered if at least one retrieved provision addresses it.

On these two representative cases, TrafficOmni-RAG averages 83\% coverage, with the advantage concentrated in Case 2 where it covers 93\% of expected dimension values compared with 9 of 15 for LlamaIndex KG. Case 1 reveals a limitation: TrafficRAG achieves higher coverage at 82\% compared with 73\% for TrafficOmni-RAG through BM25 matching, pointing to anchor extraction omissions as irrecoverable retrieval blind spots.

A blind expert evaluation on 50 randomly sampled items yields an anchor extraction F1 of 64.3\%, with precision at 57.3\% and recall at 73.3\%. The recall-dominant profile aligns with the architecture: false negatives create irrecoverable retrieval blind spots, while false positives are suppressed by downstream RRF fusion.

\subsection{Performance Across Legal Reasoning Types}\label{sec:legal-reasoning-types}

We partition TrafficLaw-QA into three types corresponding to distinct retrieval failure modes. Category A comprises 68 Single-Anchor queries testing single-dimension coverage. Category B contains 102 Multi-Anchor queries testing cross-dimensional coordination, the core challenge that OMAGR addresses. Category C includes 30 Relational/Multi-Hop queries testing multi-hop traversal depth over inter-provision dependency chains.

Table~\ref{tab:cp-cr-method} reports Context Precision and Context Recall per type.

\begin{table}[!htb]
\centering
\caption{Context Precision (CP) and Context Recall (CR) by method across different legal reasoning types.}
\label{tab:cp-cr-method}
\begin{tabular}{@{}lccc@{}}
\toprule
Method               & A (Single-Anchor) & B (Multi-Anchor) & C (Rel./Multi-Hop) \\
                     & CP~/~CR           & CP~/~CR          & CP~/~CR            \\
\midrule
TrafficOmni-RAG            & \textbf{0.941}~/~0.919 & \textbf{0.900}~/~0.863 & \textbf{0.903}~/~0.882 \\
Naive RAG            & 0.890~/~0.914         & 0.795~/~\textbf{0.876} & 0.841~/~0.888         \\
TrafficRAG           & 0.917~/~\textbf{0.924} & 0.860~/~0.872 & 0.888~/~\textbf{0.904} \\
LlamaIndex KG        & 0.713~/~0.796         & 0.736~/~0.813         & 0.700~/~0.737         \\
\bottomrule
\end{tabular}
\end{table}

TrafficOmni-RAG leads in Context Precision on all three types. The advantage over TrafficRAG follows a gradient: largest on Category B at 0.040, moderate on Category A at 0.024, and smallest on Category C at 0.015. This ranking validates the multi-anchor design logic. Category B, the intended scenario for parallel anchor traversal, simultaneously activates all four retrieval dimensions, yielding the peak gain. On Category A, adequate dimensional coverage from a single dense retrieval pass renders additional anchors redundant. On Category C, cross-reference chains extending beyond the fixed two-hop radius shift the bottleneck from dimensional breadth to traversal depth, capping the multi-anchor gain.

Context Recall follows a different pattern. TrafficOmni-RAG trails the best baseline on all three categories, with the gap widening from 0.005 on A to 0.013 on B and 0.022 on C. TrafficRAG achieves the highest Context Recall on Categories A and C through BM25 lexical matching; on Category B, Naive RAG leads at 0.876.

Table~\ref{tab:faith-ar-method} reports Faithfulness and Answer Relevancy.

\begin{table}[!htb]
\centering
\caption{Faithfulness and Answer Relevancy by legal reasoning type.}
\label{tab:faith-ar-method}
\begin{tabular}{@{}lccc@{}}
\toprule
Method               & A (Single-Anchor) & B (Multi-Anchor) & C (Rel./Multi-Hop) \\
                     & Faith.~/~AR       & Faith.~/~AR      & Faith.~/~AR        \\
\midrule
TrafficOmni-RAG            & \textbf{0.908}~/~\textbf{0.933} & 0.694~/~0.903 & 0.828~/~0.896 \\
Naive RAG            & 0.701~/~0.752         & 0.574~/~0.719         & 0.623~/~0.757         \\
TrafficRAG           & 0.821~/~0.872         & 0.616~/~0.811         & 0.697~/~0.857         \\
LlamaIndex KG        & 0.767~/~0.901         & \textbf{0.730}~/~\textbf{0.906} & \textbf{0.835}~/~\textbf{0.913} \\
\bottomrule
\end{tabular}
\end{table}

TrafficOmni-RAG leads on Category A, with Faithfulness at 0.908 and Answer Relevancy at 0.933. On Categories B and C, LlamaIndex KG records higher Faithfulness: 0.730 against 0.694 on B, and 0.835 against 0.828 on C. This inversion reflects a precision-coverage tradeoff: LlamaIndex KG retrieves from a narrow evidence base, its Context Precision trailing that of TrafficOmni-RAG by 0.164 on B and 0.203 on C, and RAGAS Faithfulness measures support by the retrieved context rather than legal completeness of that context. A system retrieving few but internally consistent provisions can score high on Faithfulness while omitting relevant statutes, a failure mode that Context Precision captures but Faithfulness does not. TrafficOmni-RAG trades a modest Faithfulness cost for a substantial Context Precision gain. The cost is asymmetric: the Faithfulness gap narrows to 0.007 on Category C and reaches 0.036 on Category B, where the steeper tradeoff is justified by the Context Precision gain of 0.040.



\section{Conclusion and Limitations}\label{sec:conclusion}

Traffic law liability determination requires simultaneous retrieval across multiple interdependent legal dimensions, a bottleneck that single-axis architectures cannot address. We propose Ontology-Guided Multi-Anchor Retrieval (OMAGR), instantiated as the core of TrafficOmni-RAG, which addresses this bottleneck through parallel multi-anchor retrieval over the traffic law ontology; experiments on TrafficLaw-QA confirm that multi-anchor parallelism drives the principal gain. Limitations include anchor extraction quality, a fixed two-hop expansion radius, and the exclusive focus on Chinese traffic law. Future work includes hardening anchor extraction, extending traversal depth, and evaluating cross-jurisdictional generalizability.

\begin{credits}
\subsubsection{\ackname} This work is supported by the Opening Project of Intelligent
Policing Key Laboratory of Sichuan Province (Grant No. ZNJW2025KFZD004).

\end{credits}
%
%
%
%
\bibliographystyle{splncs04}
\bibliography{references}

@inproceedings{ICLR2024_25f7be96,
 author = {Asai, Akari and Wu, Zeqiu and Wang, Yizhong and Sil, Avi and Hajishirzi, Hannaneh },
 booktitle = {International Conference on Learning Representations},
 title = {Self-RAG: Learning to Retrieve, Generate, and Critique through Self-Reflection},
 url = {https://proceedings.iclr.cc/paper_files/paper/2024/file/25f7be9694d7b32d5cc670927b8091e1-Paper-Conference.pdf},
 year = {2024}
}

@inproceedings{10.1145/290941.291025,
author = {Carbonell, Jaime and Goldstein, Jade},
booktitle = {ACM SIGIR Conference on Research and Development in Information Retrieval},
title = {The use of MMR, diversity-based reranking for reordering documents and producing summaries},
year = {1998},
doi = {10.1145/290941.291025},
}

@inproceedings{chalkidis-etal-2020-legal,
    title = "{LEGAL}-{BERT}: The Muppets straight out of Law School",
    author = {Chalkidis and others},
    booktitle = {Findings of the Association for Computational Linguistics: EMNLP},
    year = "2020",
    publisher = "Association for Computational Linguistics",
    url = "https://aclanthology.org/2020.findings-emnlp.261/",
    doi = "10.18653/v1/2020.findings-emnlp.261"
}

@inproceedings{10.1145/1571941.1572114,
author = {Cormack, Gordon V. and others},
booktitle = {ACM SIGIR Conference on Research and Development in Information Retrieval},
title = {Reciprocal rank fusion outperforms condorcet and individual rank learning methods},
year = {2009},
isbn = {9781605584836},
publisher = {Association for Computing Machinery},
doi = {10.1145/1571941.1572114}
}

@misc{cui2024chatlawmultiagentcollaborativelegal,
      title={Chatlaw: A Multi-Agent Collaborative Legal Assistant with Knowledge Graph Enhanced Mixture-of-Experts Large Language Model}, 
      author={Jiaxi Cui and Munan Ning and Zongjian Li and Bohua Chen and Yang Yan and Hao Li and Bin Ling and Yonghong Tian and Li Yuan},
      year={2024},
      archivePrefix={arXiv},
      primaryClass={cs.CL},
      url={https://arxiv.org/abs/2306.16092}, 
}

@InProceedings{10.1007/978-3-031-36819-6_27,
author = {Dang and others},
booktitle = {Advances and Trends in Artificial Intelligence. Theory and Applications: 36th International Conference on Industrial, Engineering and Other Applications of Applied Intelligent Systems, IEA/AIE 2023, Shanghai, China, July 19–22, 2023, Proceedings, Part I},
title="Information Retrieval from Legal Documents with Ontology and Graph Embeddings Approach",
year="2023",
publisher="Springer Nature Switzerland",
address="Cham",
isbn="978-3-031-36819-6"
}

@misc{edge2025localglobalgraphrag,
      title={From Local to Global: A Graph RAG Approach to Query-Focused Summarization}, 
      author={Darren Edge and others},
      year={2025},
      archivePrefix={arXiv},
      primaryClass={cs.CL},
      url={https://arxiv.org/abs/2404.16130}, 
}

@inproceedings{es-etal-2024-ragas,
    title = "{RAGA}s: Automated Evaluation of Retrieval Augmented Generation",
    author = {Es and others},
    booktitle = {Proceedings of the 18th Conference of the European Chapter of the Association for Computational Linguistics: System Demonstrations},
    year = "2024",
    publisher = "Association for Computational Linguistics",
    url = "https://aclanthology.org/2024.eacl-demo.16/",
    doi = "10.18653/v1/2024.eacl-demo.16"
}

@inproceedings{gao-etal-2023-enabling,
    title = "Enabling Large Language Models to Generate Text with Citations",
    author = {Gao and others},
    booktitle = {Proceedings of the 2023 Conference on Empirical Methods in Natural Language Processing},
    year = "2023",
    publisher = "Association for Computational Linguistics",
    url = "https://aclanthology.org/2023.emnlp-main.398/",
    doi = "10.18653/v1/2023.emnlp-main.398"
}

@inproceedings{NEURIPS2023_89e44582,
 author = {Guha, Neel and Nyarko, Julian and others},
 booktitle = {Advances in Neural Information Processing Systems},
 publisher = {Curran Associates, Inc.},
 title = {LegalBench: A Collaboratively Built Benchmark for Measuring Legal Reasoning in Large Language Models},
 url = {https://proceedings.neurips.cc/paper_files/paper/2023/file/89e44582fd28ddfea1ea4dcb0ebbf4b0-Paper-Datasets_and_Benchmarks.pdf},
 year = {2023}
}

@misc{huang2023lawyerllamatechnicalreport,
      title={Lawyer LLaMA Technical Report}, 
      author={Quzhe Huang and Mingxu Tao and Chen Zhang and Zhenwei An and Cong Jiang and Zhibin Chen and Zirui Wu and Yansong Feng},
      year={2023},
      archivePrefix={arXiv},
      primaryClass={cs.CL},
      url={https://arxiv.org/abs/2305.15062}, 
}

@ARTICLE{8733051,
  author={Johnson, Jeff and Douze, Matthijs and Jégou, Hervé},
  journal={IEEE Transactions on Big Data}, 
  title={Billion-Scale Similarity Search with GPUs}, 
  year={2021},
  doi={10.1109/TBDATA.2019.2921572}}

@inproceedings{karpukhin-etal-2020-dense,
    title = "Dense Passage Retrieval for Open-Domain Question Answering",
    author = {Karpukhin and others},
    booktitle = {Proceedings of the 2020 Conference on Empirical Methods in Natural Language Processing},
    month = nov,
    year = "2020",
    address = "Online",
    publisher = "Association for Computational Linguistics",
    url = "https://aclanthology.org/2020.emnlp-main.550/",
    doi = "10.18653/v1/2020.emnlp-main.550"
}

@inproceedings{NEURIPS2020_6b493230,
 author = {Lewis, Patrick and Perez, Ethan and others},
 booktitle = {Advances in Neural Information Processing Systems},
 editor = {H. Larochelle and M. Ranzato and R. Hadsell and M.F. Balcan and H. Lin},
 pages = {9459--9474},
 publisher = {Curran Associates, Inc.},
 title = {Retrieval-Augmented Generation for Knowledge-Intensive NLP Tasks},
 url = {https://proceedings.neurips.cc/paper_files/paper/2020/file/6b493230205f780e1bc26945df7481e5-Paper.pdf},
 volume = {33},
 year = {2020}
}

@article{9bca3c55c68d44ceaaa4ed350a02ccc0,
title = "Benchmarking KG-based RAG Systems: A Case Study of Legal Documents",
author = {Ongris and others},
year = "2025",
journal = "CEUR Workshop Proceedings",
issn = "1613-0073",
publisher = "CEUR-WS",
}

@ARTICLE{10387715,
  author = {Pan and others},
  journal={IEEE Transactions on Knowledge and Data Engineering}, 
  title={Unifying Large Language Models and Knowledge Graphs: A Roadmap}, 
  year={2024},
  doi={10.1109/TKDE.2024.3352100}}

@inproceedings{wang-etal-2023-query2doc,
    title = "Query2doc: Query Expansion with Large Language Models",
    author = {Wang and others},
    booktitle = {Proceedings of the 2023 Conference on Empirical Methods in Natural Language Processing},
    year = "2023",
    address = "Singapore",
    publisher = "Association for Computational Linguistics",
    url = "https://aclanthology.org/2023.emnlp-main.585/",
    doi = "10.18653/v1/2023.emnlp-main.585"
}

@article{10.1371/journal.pone.0329107,
    doi = {10.1371/journal.pone.0329107},
    author = {Wang and others},
    journal = {PLOS ONE},
    publisher = {Public Library of Science},
    title = {AMFormer-based framework for accident responsibility attribution: Interpretable analysis with traffic accident features},
    year = {2025},

}

@misc{xiao2018cail2018largescalelegaldataset,
      title={CAIL2018: A Large-Scale Legal Dataset for Judgment Prediction}, 
      author={Chaojun Xiao and others},
      year={2018},
      archivePrefix={arXiv},
      primaryClass={cs.CL},
      url={https://arxiv.org/abs/1807.02478}, 
}

@inproceedings{10.1145/3626772.3657878,
author = {Xiao and others},
booktitle = {ACM SIGIR Conference on Research and Development in Information Retrieval},
title = {C-Pack: Packed Resources For General Chinese Embeddings},
year = {2024},
isbn = {9798400704314},
publisher = {Association for Computing Machinery},
address = {New York, NY, USA},
url = {https://doi.org/10.1145/3626772.3657878},
doi = {10.1145/3626772.3657878},
location = {Washington DC, USA},
}

@InProceedings{trafficRAG,
author="Xu, Li
and Zedong, Fu
and Xinyi, Li
and Xun, Han",
title="TrafficRAG: A Multimodal RAG Framework for Trafffc Accident Liability Determination",
booktitle="Artificial Neural Networks and Machine Learning -- ICANN 2026",
year="2026",
publisher="Springer Nature Switzerland",
}

@misc{yue2023disclawllmfinetuninglargelanguage,
      title={DISC-LawLLM: Fine-tuning Large Language Models for Intelligent Legal Services}, 
      author={Shengbin Yue and Wei Chen and Siyuan Wang and Bingxuan Li and Chenchen Shen and Shujun Liu and Yuxuan Zhou and Yao Xiao and Song Yun and Xuanjing Huang and Zhongyu Wei},
      year={2023},
      archivePrefix={arXiv},
      primaryClass={cs.CL},
      url={https://arxiv.org/abs/2309.11325}, 
}
\end{document}